\documentclass[10pt, letterpaper]{article}

\usepackage[final]{corl_2020} 
\usepackage[utf8]{inputenc} 
\usepackage[T1]{fontenc}
\usepackage{booktabs}
\usepackage{amsfonts}
\usepackage{nicefrac}
\usepackage{microtype}
\usepackage{wrapfig,graphicx}
\usepackage{subfig}
\usepackage{amsmath}
\usepackage{tabularx}

\title{Multi-Resolution Graph Neural Network for Large-Scale Pointcloud Segmentation}

\author{
  Liuyue Xie\\
  Carnegie Mellon University\\ 
  United States\\
  \And
  Tomotake Furuhata\\
  Carnegie Mellon University\\
  United States\\
  \And
  Kenji Shimada \\
  Carnegie Mellon University \\
  United States\\
}

\begin{document}
\maketitle

\begin{abstract}
In this paper, we propose a multi-resolution deep-learning architecture to semantically segment dense large-scale pointclouds. Dense pointcloud data require a computationally expensive feature encoding process before semantic segmentation. Previous work has used different approaches to drastically downsample from the original pointcloud so common computing hardware can be utilized. While these approaches can relieve the computation burden to some extent, they are still limited in their processing capability for multiple scans. We present \textbf{MuGNet}, a memory-efficient, end-to-end graph neural network framework to perform semantic segmentation on large-scale pointclouds. We reduce the computation demand by utilizing a graph neural network on the preformed pointcloud graphs and retain the precision of the segmentation with a bidirectional network that fuses feature embedding at different resolutions. Our framework has been validated on benchmark datasets including Stanford Large-Scale 3D Indoor Spaces Dataset(S3DIS) and Virtual KITTI Dataset. We demonstrate that our framework can process up to 45 room scans at once on a single 11 GB GPU while still surpassing other graph-based solutions for segmentation on S3DIS with an 88.5\% (+3\%) overall accuracy and 69.8\% (+7.7\%) mIOU accuracy.
\end{abstract}



\section{Introduction}

\begin{wrapfigure}{r}{0.45\textwidth}
 \vspace{-12.5pt}
    \includegraphics[width=0.45\textwidth]{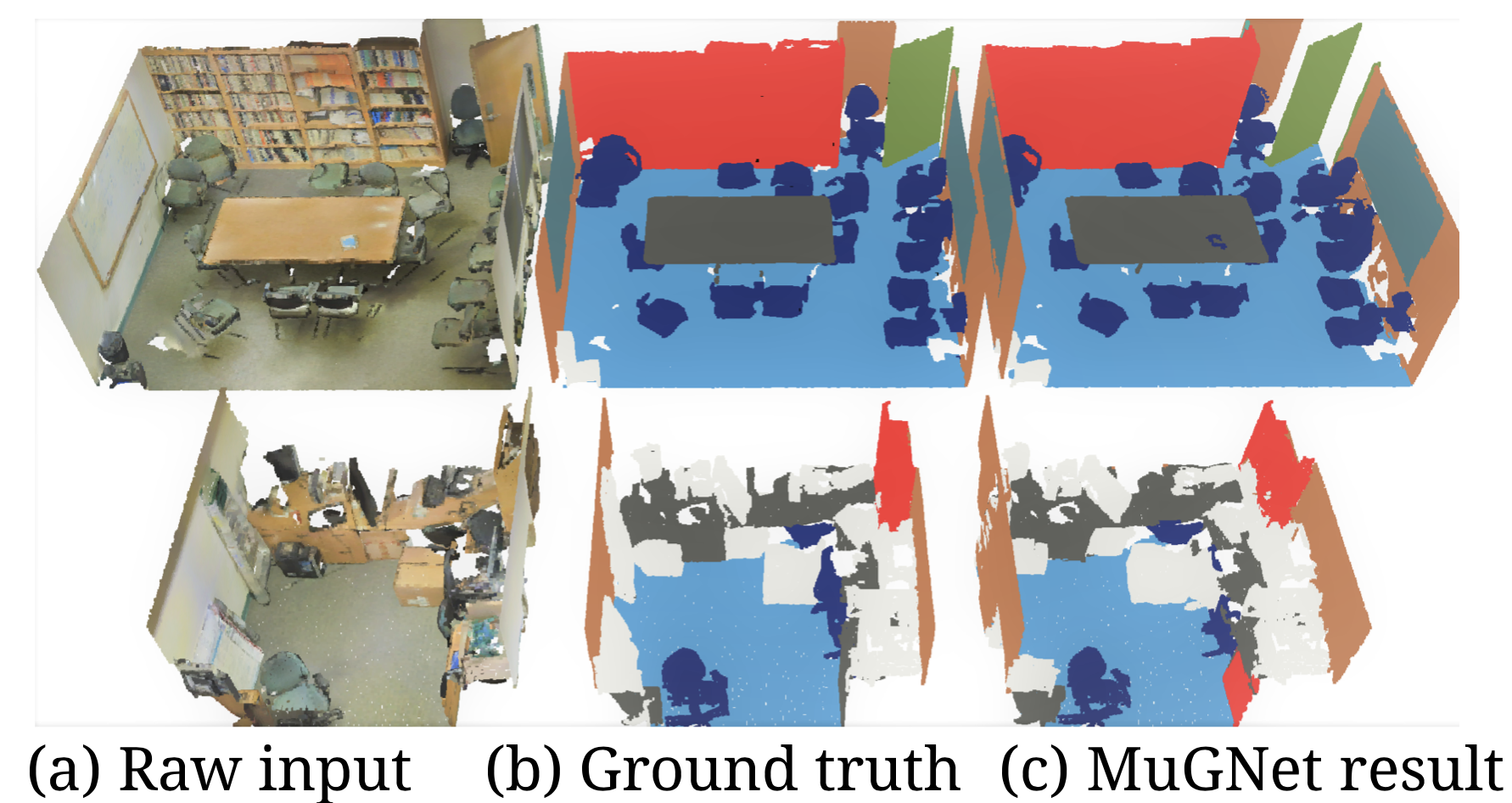}
    \captionsetup{justification=centering}

 \caption{Sample segmentation results on S3DIS \cite{2017arXiv170201105A} Area $3$.}\label{fig:area6}
    \vspace{-10pt}
\end{wrapfigure}

Semantic segmentation of large-scale 3D pointcloud data has attracted numerous interests in real-world applications. Automatic sight inspection for quality verification, for example, is one of the emerging applications for pointcloud semantic segmentation. Deployment of a 3D pointcloud semantic segmentation algorithm can avoid delays and reduce costs caused by human errors. Figure \ref{fig:area6} shows how accurately our proposed multi-resolution graph neural network can perform and semantic segmentation. 

When it comes to dense large-scale pointcloud scans, the current methods either require a significant reduction in the data density \cite{kpconv, qi2016volumetric, DBLP:journals/corr/LiPSQG16, DBLP:journals/corr/EngelckeRWTP16, pointnet, shelnet} or offer only limited processing capability for multiple scans \cite{DBLP:journals/corr/MasciBBV15, 10.1007/978-3-030-01237-3_6, pointnet2, pointcnn, pointweb}. Two challenges hinder the development of pointcloud semantic segmentation algorithms on dense large-scale pointcloud scans. 

The first is the extensive memory usage for processing dense pointclouds with up to billions of data points in a single scan.
Prior research has attempted to downsample from the original pointcloud to the state that common computing hardware can be utilized. 
Performing drastic downsampling on dense pointcloud can, however, negatively impact the segmentation result on dense pointcloud. 
Intricate details initially present in the dense pointcloud input are removed during the downsampling process. This is undesirable since sparse pointclouds contain less geometric features. 
The segmentation result obtained with sparse pointcloud would be inaccurate when interpolating back onto the original dense pointcloud. To address this issue, we convert the large-scale pointclouds into semantically similar point clusters. By doing this, the amount of GPU memory demanded by the semantic segmentation network is drastically reduced. When tested on the benchmarked dataset Stanford Large-Scale 3D Indoor Spaces Dataset, we observe that MuGNet can inference at once up to 45 pointclouds, each containing on average 2.6 million points.

Besides the challenge posed by computation requirement, semantic segmentation is also challenged by the orderless and uneven nature of the raw pointclouds. 
This leads to the less desirable performance of deep convolutional neural networks, which require evenly arranged and ordered data. Prior research has attempted to convert the orderless pointclouds to mesh or voxels as ways to artificially structure pointclouds before applying deep convolutional neural networks \cite{qi2016volumetric, DBLP:journals/corr/LiPSQG16, DBLP:journals/corr/EngelckeRWTP16, acnn, interpolate}. 
Such attempts typically result in voluminous rendered data and introduce unnecessary computation overhead to the algorithm. To overcome this challenge, we propose a framework with graph convolutions that are invariant to permutations of pointcloud orderings. Segmentation can thus be performed without artificially structuring the pointcloud. 

In addition to addressing the two major challenges, our proposed framework also features a bidirectional multi-resolution fusion network. 
The framework reasons the relationship between adjacent clusters at different resolutions with both forward and backward paths. We observe that the concatenated graph features from different resolutions provide richer feature representation compared to the resultant features from the final convolution layer alone. The backward fusion network then further enriches the representations. With the aforementioned design components, we demonstrate that our proposed MuGNet achieves 88.5\% overall accuracy and 69.8\% mIOU accuracy on Stanford Large-Scale 3D Indoor Spaces Dataset semantic segmentation, outperforming SPG \cite{DBLP:journals/corr/abs-1711-09869} by 7.7\% better mIOU accuracy and 3\% better overall accuracy. Figure \ref{fig:area6} presents a sample semantic segmentation result for Area 3 in the Stanford Large-Scale 3D Indoor Spaces Dataset. 


\section{Related Work}
\label{sec:citations}
Three main categories of learning-based frameworks have been previously proposed for pointcloud semantic segmentation: voxel-based approach, point-based approach and graph-based approach. We outline the corresponding approaches as follows.

\hspace{5mm} \textbf{Voxel-based approach:} As attempts to tackle the orderless and uneven nature of pointcloud, previous algorithms have ventured to convert pointcloud into structured voxel data such that convolution neural networks can be applied. Volumetric CNN \cite{qi2016volumetric} for example,  pioneered on voxelized shapes; 
FPNN \citep{DBLP:journals/corr/LiPSQG16} and Vote3Deep \citep{DBLP:journals/corr/EngelckeRWTP16} proposed special methods to deal with the sparsity problem in volumes. Voxelizing a large pointcloud scan, however, imposes computation overhead. Such operation becomes infeasible for processing dense large-scale pointclouds. 

\hspace{5mm} \textbf{Point-based approach:} PointNet \cite{pointnet} has inspired many previous works to process pointclouds as direct input. Typical point-based frameworks learn feature encodings of each point. The encoded features are then aggregated with a permutation invariant function to arrive at transformation invariant segmentation results. Spatial and spectral convolutions have been implemented in previous works \cite{DBLP:journals/corr/MasciBBV15, Tang2019ChebNetEA,10.1007/978-3-030-01237-3_6, acnn, kpconv, pointcnn, pointnet2} to improve the efficiency of feature encoding. All of these approaches are demanding in memory usage and thus require either a sliding window approach or data downsampling approach to process dense pointclouds. In contrast, our framework alleviates the memory demand during the segmentation process by preforming point clusters based on geometric similarities. In this way, we can process multiple dense pointclouds at once with a commonly available GPU. 

\hspace{5mm} \textbf{Graph-based approach:} While pointclouds are inherently orderless and unevenly distributed, they can be represented as graphs with interdependency between points. Node classification can be performed to distinguish the classes among the nodes in the graph representations \cite{DBLP:journals/corr/KipfW16, DBLP:journals/corr/HamiltonYL17,velickovic2018graph}. Wang et al.\cite{DBLP:journals/corr/abs-1801-07829} first applied the concept of graph convolutional network on pointcloud data and formulated an approach that dynamically computes the graphs at each layer of the network. Various types of graph convolutional neural networks have since been utilized in the context of pointcloud segmentation \cite{mining, Wang2019_GACNet, DBLP:journals/corr/abs-1711-09869, eng, dgcnn}. All of the frameworks sequentially infer the graph embeddings with only forward network paths. On the contrary, we exploit a bidirectional framework that retains rich contextual information derived from multiple convolution units. The backward path of our network receives graph embeddings at different resolutions and infer rich contextual relationships to achieve high segmentation accuracy. 

\section{Proposed Computational Framework}

\label{sec:methodology}

    \begin{figure}[htbp]
    \centering
    \includegraphics[width=0.85\textwidth]{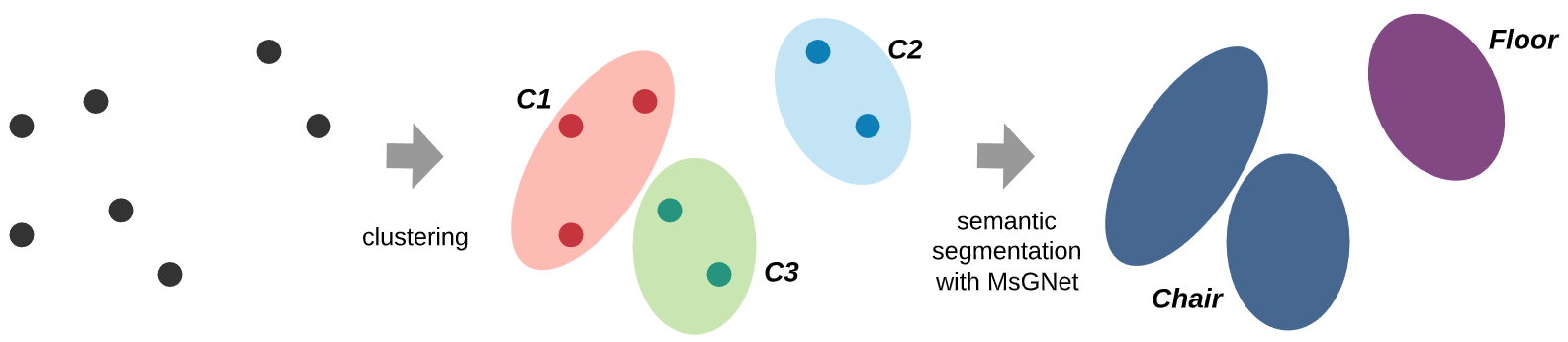}
    \captionsetup{justification=centering}
    \caption{Illustration of the overall workflow for MuGNet. The input pointcloud is first clustered, and then each formed cluster is classified into its respective semantic classes based on cluster-features.}
    \label{overview}
    \end{figure}
    
    Given a large-scale pointcloud with millions of points spanning up to hundreds of meters, directly processing it with a deep neural network requires an innumerable amount of computation capability. Downsampling points from the original pointcloud by order of magnitude is a common practice to cope with this limitation. This approach, however, takes away the intricate details in the original pointcloud and yet still suffers from expensive memory usage. We propose MuGNet, a multi-resolution graph neural network inspired by EfficientDet \cite{tan2019efficientdet}, to effectively translates large-scale pointclouds into directed connectivity graphs and efficiently segments the pointclouds from the formed graph with a bidirectional graph convolution network. MuGNet allows the processing of multiple large-scale pointclouds in a single pass while producing excellent segmentation accuracy.
    Figure \ref{overview} showcases the overall workflow of MuGNet. In the subsequent sections, we will further explain the cluster formation process and introduce the three key design features in our segmentation network: cluster feature embedding, feature-fusion backbone, and bidirectional-convolution network.

\subsection{Clustering algorithm for graph formation}
\label{sec:cluster}
    We preprocess large-scale pointclouds into geometrically homogeneous clusters, a 3D equivalent of superpixels commonly used in image analysis. The existing point clustering approaches can be roughly classified into unsupervised geometric approaches and supervised learning-based approaches. There is currently no standard clustering strategy that is suitable for large-scale pointclouds. We individually analyze their relative merits as follows and indicate our reasoning for choosing the supervised clustering approach.
    
    \hspace{5mm} \textbf{(1) Unsupervised clustering}
    Given a pointcloud data with millions of points, geometric features based on the neighborhood planarity, linearity, and scattering can be calculated \cite{article, DBLP:journals/corr/abs-1711-09869}. The unsupervised methods rely on the assumption that geometrically or radio-metrically homogeneous segments are also semantically homogeneous. This assumption is challenged when attempting to form semantically homogeneous clusters for semantic segmentation \cite{DBLP:journals/corr/abs-1904-02113}. Since the clusters are formed without the oversight of semantic information, the clusters can have semantic impurity among its constituting points. The impurity within each of the clusters negatively impacts the performance of the subsequent node classification network and limits the final semantic-segmentation accuracy.
    
\begin{figure}[htbp]
    \centering
    \captionsetup{justification=centering}
    \includegraphics[width=\linewidth]{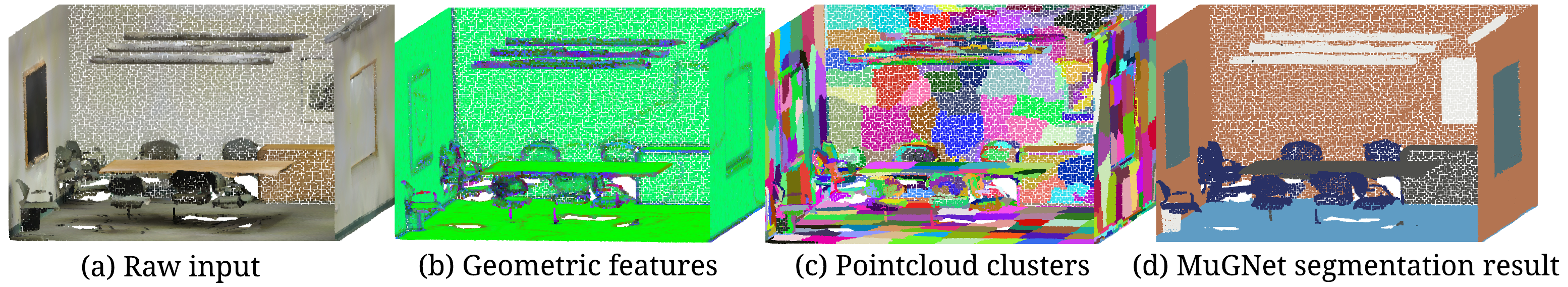}
    \caption{Qualitative semantic segmentation results of MuGNet on the validation set of S3DIS \cite{2017arXiv170201105A}.}
    \label{s3dis}
    \end{figure}

    \hspace{5mm} \textbf{(2) Supervised clustering}
	The supervised-clustering method proposed by Landrieu et al.\cite{DBLP:journals/corr/abs-1904-02113} standardizes a pointcloud with a spatial transformer, embeds local feature with a small MLP-based network, and finally forms clusters by optimizing on the generalized minimal partition problem with the introduction of contrastive loss as a feedback metric for cluster purity. The supervised clustering approach achieves significant improvements compared to the unsupervised methods in terms of reduction in semantic impurity within the formed clusters. The semantically pure clusters can provide enhanced geometric and semantic connectivity information for subsequent learning tasks. On average the supervised clustering technique can effectively encapsulate pointcloud information of a Stanford Large-Scale 3D Indoor Spaces Dataset room scan containing $\sim2.6$ million points into $\sim10^3$ clusters. Compared to downsampling a pointcloud scan randomly, the conversion to point clusters can perform a drastic reduction in data size while carrying richer contextual information from the original pointcloud. Figure \ref{s3dis}(c) and \ref{vKITTI}(c) visualize sample pointcloud scans color-coded by geometric features and point clusters produced by the supervised clustering approach.
\subsection{Cluster-feature Embedding}
\label{sec:clusteremb}

\begin{figure}[h]
\centering
\captionsetup{justification=centering}
\includegraphics[width=0.9\linewidth]{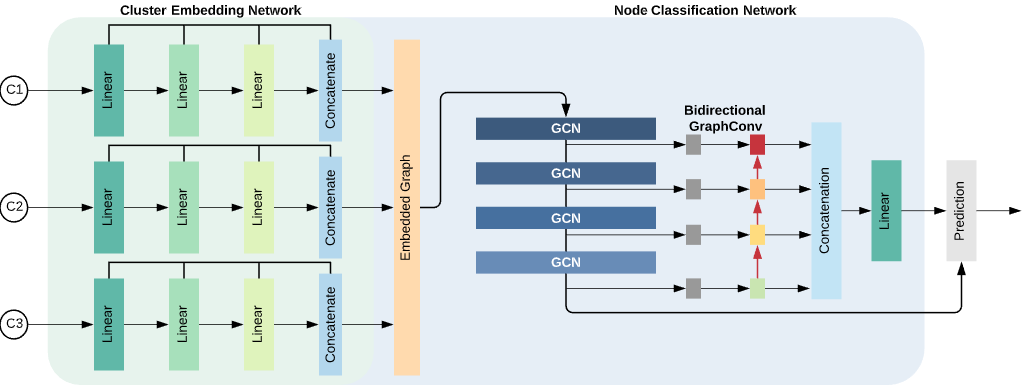}
\caption{MuGNet includes cluster embedding and bidirectional graph convolution networks. The input pointcloud is clustered based on their geometric similarities and learnable features into cluster set $C = \{C_1, C_2, ...C_K\},$ where $K$ denotes the total number of clusters formed for a pointcloud.}
\label{network}
\end{figure}
    As shown in Figure \ref{network}, our cluster-feature embedding network is applied to each of the formed pointcloud clusters. The points in a cluster go through a multi-resolution feature-fusion network with three sets of multi-layer perceptrons. The input points contain geometric features analogous to \cite{DBLP:journals/corr/abs-1711-09869}. The feature-fusion network receives pointcloud at three different resolutions. The highest resolution channel is defined as the cluster points, from which the other two lower-resolution channels are formed by down-sampling with linear layers. Each of the three-channel inputs then goes through a series of 2D convolutions, and their output features are concatenated together into a single feature vector. Utilization of the feature-fusion network for cluster-feature extraction effectively identifies cluster-features that are not only local representative but also distinctive at a larger receptive field. 

\subsection{Feature-fusion Backbone}
\label{sec:backbone}

\begin{wrapfigure}{r}{0.25\textwidth}
\captionsetup{justification=centering}
 \vspace{-17.5pt}
    \includegraphics[width=0.25\textwidth]{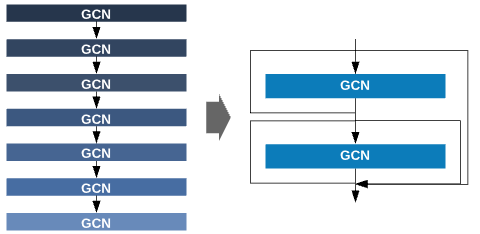}
 \caption{Exploded-view of the backbone layers.}\label{fig:backbone}
    \vspace{-5pt}
\end{wrapfigure}

    The node-classification network that identifies the semantic class for each formed cluster employs a backbone network to extract feature vectors at different resolutions. The feature vectors are subsequently fed to a bidirectional pyramidal feature-fusion network. Consider a graph $G(V, E)$ containing point clusters as nodes and node features obtained from the embedding network, where $V = {1, 2, ... N}$ and $E \subseteq |V|\times|V|$ represent the set of vertices and edges in the formed graph respectively. $N$ denotes the total number of nodes. The set of neighboring nodes of Node $i$ can be denoted as $N(i) = \{j:(i, j) \in E\} \cup \{i\}$. Each of the individual node feature $h_i\in\mathbb{R}^F$ is associated with a corresponding graph Node $i\in V$, among the set of node features, $H = \{h_1, h_2, ..., h_N\}$, where $F$ denotes feature dimension at each node.

    The backbone network consists of four residual blocks, each constructed with a graph convolution, batch normalization, and activation layer: 
    $$H^{bb out}_i = Activation(BatchNorm(GraphConv(H^{bb in}_i))),$$
    where $i$ denotes the level of backbone. 
    It has been observed in previous works that with the message-passing mechanism for graph convolutions, the node features eventually become similar as the number of passing increases. Short-term and long-term residual connections are added to the basic building blocks to increase information density and avoid gradient diminishing as the network depth increases. For the short-term residual connection, the output features from the previous block are aggregated with a simple addition operation. The expression of the residual block with a short-term direct connection from the previous block is formulated as:
    $$H^{bb out}_2 = \{Activation(BatchNorm(GraphConv(H^{bb in}_2))) + H^{bb out}_1\}.$$
    The outputs from the residual blocks are fed into the bidirectional graph convolution network in parallel. At the end of the backbone structure, the features are concatenated together for cluster classification:
    $$H^{bb out} = \{H^{bb out}_1, H^{bb out}_2, ..., H^{bb out}_4\},$$
    where $H^{bb out}$ expresses the concatenated final feature output from the backbone network.

\subsection{Bidirectioal-graph Convolution Network}
\label{sec:bidirection}
    
    Multi-scale feature-fusion aims to aggregate features at different resolutions. Given a list of multi-scale features, $H^{in} = (H^{in}_{l1}, H^{in}_{l2}, ...)$, where each of $H^{in}_{li}$ represents the feature at level $l_i$, the feature-fusion network acts as a transformation function, $f$, that aggregates features at different resolution and outputs a list of new features denoted as $H^{out} = f(H^{in})$. The structure of the bidirectional graph convolution network is shown in Figure \ref{network}. The network receives graph features, $H^{in} = (H^{in}_{1} ...H^{in}_{4})$, at different resolutions from backbone levels 1-4.  The conventional feature pyramidal network used in image tasks aggregates multi-scale features in a top-down manner and performs resizing operation for resolution matching. Each of the network node in the pyramidal network is formulated in a similar fashion to the basic backbone building block, where batchnorm and activation layers are incorporated on top of each of the graph convolution: 
    $$H^{out}_i = Activation(BatchNorm(GraphConv(H^{in}_i))).$$

    For graph convolutions, resizing becomes inefficient and causes information mixing during the message passing operation. The feature pyramidal network is thus redesigned to produce a constant number of features at each level:
\begin{align*}
    & H^{out}_4 = GraphConv(H^{in}_4),\\
    & H^{out}_3 = GraphConv(H^{in}_3+H^{out}_4),\\
    & \cdots \\
    & H^{out}_1 = GraphConv(H^{in}_1+H^{out}_2).
\end{align*}

    While propagating the node features through a pyramidal network complements the baseline backbone structure by enriching the aggregated feature information, it is still limited by the one-way information flow. To address this issue, we configure a bidirectional graph propagation network that creates a two-way information flow to aggregate information for both the deep-to-shallow direction and the other way around. In addition to the bidirectional passes, a skip connection that travels from the input node to output node is incorporated for each resolution. In this way, feature information is more densely fused, and diminishing-gradient issue can be avoided to an extent. 
    
    Resultant features from the fusion network for different resolutions contribute to the final output feature unequally, and naively concatenating the features together requires an unnecessarily heavy memory usage. Instead, an additional set of weights are introduced when aggregating the multi-res output features into a single feature vector. The intuition is to have the network itself learn the importance of each resolution during training. This approach not only avoids manual assignment of importance based on potentially biased or inaccurate human knowledge but also leads to a more elegant memory usage compared to naive concatenation. To give a sample formulation of the final graph bidirectional GraphConv network with learnable weights for different resolutions, the formulation for aggregation at level 3 is shown as follows:
\begin{align*}
    & H^{mid}_3 = GraphConv(\frac{w_1 \cdot H^{in}_3 + w_2 \cdot H^{in}_4}{w_1 + w_2 + \epsilon}),\\
    & H^{out}_3 = GraphConv(\frac{w_1' \cdot H^{in}_3 + w_2' \cdot H^{mid}_3 + w_3'\cdot H^{out}_2}{w_1' + w_2' + w_3' + \epsilon}),
\end{align*}
    where $H^{in}_3$ indicates the middle feature-passing node at the third level on the top-to-bottom pathway. Features propagated from channels for the other resolutions are constructed in a similar fashion. The pathway and network edges are formulated according to connections shown in Figure \ref{network}. \\
    We have found through experiments that graph neural network is prone to experiencing information assimilation, which leads to ineffective feature-fusion at deeper convolution levels. This phenomenon is conceptually analogous to the thermodynamic equilibrium state where the temperature gradient between two objects diminishes as the energy from one object is transferred to another. To address this problem, we send the output features from both the backbone network and the pyramidal fusion network to the segmentation module. 

\section{Results}
\label{sec:result}
    In this section, we evaluate the MuGNet on various 3D pointcloud segmentation benchmarks, including the Stanford Large-Scale 3D Indoor Spaces(S3DIS) dataset \cite{2017arXiv170201105A}, and the Virtual KITTI(vKITTI) dataset \cite{DBLP:journals/corr/GaidonWCV16}. The network performance is evaluated by the mean Intersection-over-Union(mIOU), and Overall Accuracy (OA) of all object classes specific to each dataset.  
\subsection{Segmentation Results on S3DIS}
\label{sec:benchmark}

    The Stanford Large-Scale 3D Indoor Spaces Dataset provides a comprehensive clean collection of indoor pointcloud scans. There are in total over 695 million points and indoor scans of 270 rooms \cite{2017arXiv170201105A}. 
     Each scan is a dense pointcloud of a medium-sized room ($ \sim 20\times15\times5$ meters). We use the standard $6$-fold cross-validation approach in our experiments.
    
    Table \ref{s3dis_6fold} shows the network's performance averaged over all six areas. Our network has achieved better performance than state-of-the-art graph-based methods. When compared to the latest state-of-the-art for non-graph-based network RandLA-Net \cite{hu2019randla}, our network also achieves comparable results with 0.5\% higher overall accuracy and only 0.2\% lower mIOU. It should be noted that our network has been validated to consistently arrive at the reported result for the five times that the network is trained. In contrast, some of the baselines tend to produce inconsistent results due to their random sampling operations.

\begin{table}[!htbp]
    \setlength{\tabcolsep}{2pt} 
    \centering
    \captionsetup{justification=centering}
    \scriptsize	
    \begin{tabular}{c|cc|ccccccccccccc}
    \toprule
        \textbf{} & \textbf{OA} & \textbf{mIOU} & \textbf{ceiling} & \textbf{floor} & \textbf{wall} & \textbf{beam} & \textbf{column} & \textbf{window} & \textbf{door} & \textbf{chair} & \textbf{table} & \textbf{bookcase} & \textbf{sofa} & \textbf{board} & \textbf{clutter}\\ \midrule
        \textbf{PointNet \cite{pointnet}} & 78.5 & 47.6 & 88 & 88.7 & 69.3 & 42.4 & 23.1 & 47.5 & 51.6 & 42 & 54.1 & 38.2 & 9.6 & 29.4 & 35.2\\
        \textbf{Engelmann et al. \cite{eng}} & 81.1 & 49.7 & 90.3 & 92.1 & 67.9 & 44.7 & 24.2 & 52.3 & 51.2 & 47.4 & 58.1 & 39 & 6.9 & 30.0 & 41.9\\
        \textbf{DGCNN \cite{dgcnn}} & 84.1 & 56.1 & - & - & - & - & - & - & - & - & - & - & - & - & -\\
        \textbf{SPG \cite{DBLP:journals/corr/abs-1711-09869}} & 85.5 & 62.1 & 89.9 & 95.1 & 76.4 & 62.8 & 47.1 & 55.3 & 68.4 & 73.5 & 69.2 & 63.2 & 45.9 & 8.7 & 52.9\\
        \textbf{RandLA-Net \cite{hu2019randla}} & 88.0 & \textbf{70.0} & - & - & - & - & - & - & - & - & - & - & - & - & -\\
        \textbf{MuGNet (ours)} & \textbf{88.5} & 69.8 & \textbf{92.0} & \textbf{95.7} & \textbf{82.5} & \textbf{64.4} & \textbf{60.1} & \textbf{60.7} & \textbf{69.7} & \textbf{82.6} & \textbf{70.3} & \textbf{64.4} & \textbf{52.1} & \textbf{52.8} & \textbf{60.6}\\

    \bottomrule
    \end{tabular}
    \vspace{0.5em}
    \caption{Semantic segmentation results measured by overall accuracy, mean intersection over union, and intersection over union of each class for all six areas in S3DIS dataset \cite{2017arXiv170201105A}.}
    \label{s3dis_6fold}
\end{table}

\begin{table}[!htbp]
    \setlength{\tabcolsep}{2pt} 
    \centering
    \captionsetup{justification=centering}
    \scriptsize	
    \begin{tabular}{c|cc|ccccccccccccc}
    \toprule
        \textbf{ } & \textbf{OA} & \textbf{mIOU} & \textbf{ceiling} & \textbf{floor} & \textbf{wall} & \textbf{beam} & \textbf{column} & \textbf{window} & \textbf{door} & \textbf{chair} & \textbf{table} & \textbf{bookcase} & \textbf{sofa} & \textbf{board} & \textbf{clutter}\\ \midrule
        \textbf{PointNet \cite{pointnet}} & - & 41.1 & 88.8 & 97.3 & 69.8 & 0.05 & 3.9 & 46.3 & 10.8 & 52.6 & 58.9 & 40.3 & 5.8 & 26.4 & 33.2\\
        \textbf{Engelmann et al. \cite{eng}} & - & 48.9 & 90.1 & 96.0 & 69.9 & 0.0 & 18.4 & 38.3 & 23.1 & 75.9 & 70.4 & 58.4 & 40.9 & 13.0 & 41.6\\
        \textbf{SPG \cite{DBLP:journals/corr/abs-1711-09869}} & 86.4 & 58.0 & 89.3 & 96.9 & 78.1 & 0.0 & 42.8 & 48.9 & 61.6 & 84.7 & 75.4 & 69.8 & 52.6 & 2.1 & 52.2\\
        \textbf{GACNet \cite{Wang2019_GACNet}} & 87.8 & 62.8 & \textbf{92.3} & \textbf{98.3} & 81.9 & 0.0 & 20.3 & \textbf{59.1} & 40.8 & 78.5 & \textbf{85.8} & 61.7 & \textbf{70.7} & \textbf{74.7} & 52.8\\
        \textbf{MuGNet (ours)} & \textbf{88.1} & \textbf{63.5} & 91.0 & 96.9 & \textbf{83.2} & \textbf{5.0} & \textbf{37.0} & 54.3 & \textbf{62.6} & \textbf{85.3} & 76.4 & \textbf{70.1} & 55.2 & 55.2 & \textbf{53.4}\\

    \bottomrule
    \end{tabular}
    \vspace{0.5em}
    \caption{Semantic segmentation results measured by overall accuracy, mean intersection over union, and intersection over union of each class in Area $5$ of S3DIS dataset \cite{2017arXiv170201105A}.}
    \label{s3dis_a5}
\end{table}
    
    We would also like to investigate our network's performance compared to GACNet \cite{Wang2019_GACNet}, a state-of-the-art graph convolution for pointcloud semantic segmentation. The results for Area $5$ are compared since this is the only area reported for the work. As shown in Table \ref{s3dis_a5}, our network still achieves better performance than the rest of the baselines for the segmentation task on Area $5$. Besides, most of the compared baseline networks struggle with handling a single pointcloud scan at once. They require drastic down-sampling and dividing of a large pointcloud room into small $1\times 1 \times 1$ blocks with sparse points. Our network, on the other hand, can effectively process multiple rooms in a single pass.

\subsection{Segmentation Results on Virtual KITTI}
\label{sec:vKITTI}

\begin{figure}[htbp]
    \centering
    \captionsetup{justification=centering}
    \includegraphics[width=\linewidth]{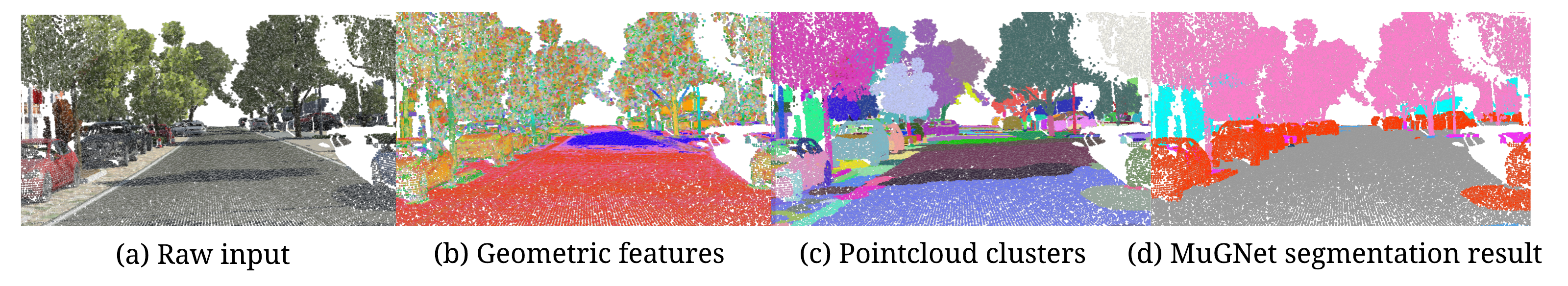}
    \caption{Qualitative semantic segmentation results of MuGNet on the validation set of vKITTI \cite{DBLP:journals/corr/GaidonWCV16}.}
    
    \label{vKITTI}
    \end{figure}
    The Virtual KITTI (vKITTI) dataset \cite{DBLP:journals/corr/GaidonWCV16} contains simulated LiDAR data acquired through $50$ annotated $1242 
    \times 375$ resolution monocular videos generated from five different worlds in urban setting. 
    Each of the scans in the dataset is a dense pointcloud with $13$ semantic classes. The annotated 2D depth images are projected into 3D space to produce the simulated LiDAR scans that resemble pointclouds obtained by real-life LiDAR scanners. For testing and training, the scans are separated into $6$ non-overlapping sets. We obtain the final evaluation following the $6$-fold validation protocol.
    
    \begin{table}[!htbp]
        \setlength{\tabcolsep}{2pt} 
        \centering
        \captionsetup{justification=centering}
        \scriptsize	
        \begin{tabular}{c|cc|ccccccccccccc}
        \toprule
            \textbf{} & \textbf{OA} & \textbf{mIOU} & \textbf{terrain} & \textbf{tree} & \textbf{vegetation} & \textbf{building} & \textbf{road} & \textbf{g-rail*} & \textbf{t-sign*} & \textbf{t-light*} & \textbf{pole} & \textbf{misc} & \textbf{truck} & \textbf{car} & \textbf{van}\\ \midrule
            \textbf{PointNet \cite{pointnet}} & 63.3 & 17.9 & 32.9 & 76.4 & 11.9 & 11.7 & 49.9 & 3.6 & 2.8 & 3.7 & 3.5 & 0.7 & 1.5 & 25.1 & 3.4\\
            \textbf{Engelmann et al. \cite{eng}} & 73.2 & 26.4 & 38.9 & 87.1 & 14.6 & 44.0 & 58.4 & 12.4 & 9.4 & 10.6 & 5.3 & 2.2 & 3.6 & 43.0 & 13.3\\
            \textbf{MuGNet (ours)} & \textbf{85.1} & \textbf{50.0} & \textbf{70.0} & \textbf{88.6} & \textbf{35.2} & \textbf{63.0} & \textbf{80.2} & \textbf{40.8} & \textbf{32.0} & \textbf{56.3} & \textbf{23.4} & \textbf{3.92} & \textbf{7.1} & \textbf{84.3} & \textbf{65.4}\\
        \bottomrule
        \end{tabular}
        \vspace{0.5em}
        \caption{Overall accuracy, mean intersection over union, and intersection over union of each class for all six splits in vKITTI dataset\cite{DBLP:journals/corr/GaidonWCV16}. *"t-" is short for traffic; "g-" is short for guard.}
        \label{vkitti_6fold}
    \end{table}
    
    Table \ref{vkitti_6fold} shows a quantitative comparison of MuGNet against other benchmark methods for the dataset, and Figure \ref{vKITTI} presents MuGNet's qualitative segmentation result. The reported model is trained with the XYZ coordinates without RGB values to investigate our model's ability to learn geometric features. For comparison purposes, the benchmarked models are also trained with the same configuration.

    MuGNet has demonstrated to exceeded previous approaches in all evaluation metrics. 
    It should be noted that during inference MuGNet can fit all $15$ scans in each of the six splits into a single GPU simultaneously, making it more viable for inspection tasks where computation resource is limited.
    
\subsection{Efficiency Analysis}
\label{sec:efficiency}
\begin{figure}[!htb]

\minipage{0.6\textwidth}
\scriptsize
  \begin{tabular}{
   p{.16\textwidth}|
   p{.33\textwidth}|
   p{.33\textwidth}
}
    \toprule
        \textbf{\# of Rooms} & \textbf{Mean Inference Time(sec)} & \textbf{GPU Memory Usage(GB)} \\ \midrule
        \textbf{1} & 0.5056 & 1.042\\
        \textbf{5} & 0.5396 & 4.073\\
        \textbf{12} & 0.6124 & 6.684\\
        \textbf{34} & 0.7077 & 10.365\\
        \textbf{45} & 0.8332 & 10.617\\
    \bottomrule
  \end{tabular}
  \vspace{14pt}
  \captionof{table}{Efficiency analysis on S3DIS dataset \cite{2017arXiv170201105A}.}\label{table:efficiency_table}
\endminipage\hfill
\minipage{0.4\textwidth}
\vspace{-4pt}
  \includegraphics[width=\linewidth]{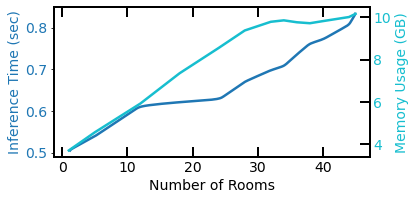}
  \captionof{figure}{Corresponding efficiency plot.}\label{fig:efficiency_plot}
\endminipage\hfill
\end{figure}

    The scalability of MuGNet is investigated by increasing the number of rooms to be processed in a single shot with the Stanford Large-Scale 3D Indoor Spaces(S3DIS) dataset \cite{2017arXiv170201105A}. Table \ref{table:efficiency_table} showcases the number of rooms for inference against the respective inference time and GPU memory consumption, and Figure \ref{fig:efficiency_plot} presents the corresponding plot of the trend. With a setup of single NVIDIA RTX $1080$Ti GPU, up to $45$ rooms (totaled $38,726,591$ points) can be accommodated into the available computation resource. We have observed that with an increase of room number, the increase in memory usage slows down and resembles a pseudo-logistic growth. The inference time also only increases at a relatively low rate. The growth trends indicate that the model scales nicely with an increased number of scans to be processed. The model would, therefore, be desirable for the application scenario where a multitude of dense scans need to be processed with few shots. 

\subsection{Ablation Study}
\label{sec:ablation}
In order to further investigate the behavior of our network, we conduct the following ablation studies. All ablated networks are trained and tested with the same $6$-fold validation setup as previously mentioned in Section 4.1 .

\hspace{5mm} \textbf{(1$\sim$3) Removing Bidirectional GraphConv:} We want to study the backbone's ability to propagate useful information. In conventional convolutions, it has been widely proven that deeper networks often produce features at a higher quality that are more representative of the entire data distribution and in turn enhance network performance. If this phenomenon remains true for graph convolutions, then using features from deeper layers of the backbone as input to Bidirectional GraphConv should in theory yield a higher accuracy segmentation result. After removing the bidirectional network entirely, we obtain the final features directly from the backbone network before feeding them to the segmentation network. We also vary the number of backbone blocks to $7, 14$, and $28$ to investigate the extent of the backbone structure's capability. From the three different depths that the backbone has been tested, it can be observed that increased backbone depth in fact leads to better performance. 

\hspace{5mm} \textbf{(4) Stacking multiple Bidirectional GraphConvs:} The Bidirectional GraphConv network is structurally capable of stacking multiple network copies in parallel. Previous works in 2D image processing have shown positive correlation between model performance and the number of stacked networks. However, graph convolutions are fundamentally different from the conventional convolutions for 2D images, where the filtering operation is replaced by a message-passing mechanism. Naively adopting the structure used in 2D image processing would lead to sub-optimal results.

\hspace{5mm} \textbf{(5) Increase depth of backbone layers:}  
Previous studies on graph neural networks have alerted on the loss of expressive power as the network complexity increases \cite{express, power}. It is therefore crucial to investigate the impact of backbone depth to the Bidirectional GraphConv network. When the backbone layer is increased to $14$, and the last $4$-layer outputs are fed into the Bidirectional GraphConv, the resultant segmentation is deemed to be less optimal compared to having $4$ layers of the backbone. Besides, the GPU memory usage increases by $\sim1/3$ times from our final network. Despite the incorporation of short-term and long-term residual connections, graph convolutions are still prone to information assimilation among nodes. A deeper backbone network does not necessarily extract higher-quality features.

\begin{table}[!htbp]
    \setlength{\tabcolsep}{2pt} 
    \centering
    \captionsetup{justification=centering}
    \scriptsize	
    
    \begin{tabular}{p{25em} | p{3.5em}}
    \toprule
        \textbf{ } & \textbf{mIOU\%}\\ \midrule
        \textbf{(1) Backbone with 7 layers} & 59.0\\
        \textbf{(2) Backbone with 14 layers} & 53.4\\
        \textbf{(3) Backbone with 28 layers} & 64.5\\
        \textbf{(4) Stacking 2 bidirectional GraphConvs} & 66.3\\
        \textbf{(5) MuGNet with 14-layer backbone} & 63.7\\
        \textbf{(6) MuGNet Baseline} & \textbf{69.8}\\

    \bottomrule
    \end{tabular}
    \vspace{0.5em}
    \caption{The $6$-fold validated mean IOU of ablated networks tested on the S3DIS dataset \cite{2017arXiv170201105A}.\\ 
    The results are compared against the final MuGNet configuration.}
    \label{experiments}
\end{table}
\vspace{-2em}

\section{Conclusion}
\label{sec:conclusion}
    In this paper, we have demonstrated that it is possible to process a large quantity of dense pointcloud scans by reformulating the learning objective to utilize a bidirectional graph convolutional network. Most of the current approaches rely on drastic downsampling or sliding window operations, both of which negatively impact the segmentation result on dense pointclouds. In contrast, we effectively transform the pointclouds into graph representations, which radically reduce the computation demand and facilitate the processing of up to forty-five room scans with a single commercially available GPU. A multi-resolution cluster embedding network is introduced to provide high-quality representations of the point clusters. Finally, a bidirectional-graph convolution network aggregates useful features from different graph resolutions. Extensive experiments on benchmark datasets have proven our network's strong capability in processing a multitude of pointclouds. It has also achieved state-of-the-art accuracy for pointcloud semantic segmentation. Our model would be desirable for application scenarios where a multitude of pointcloud scans need to be processed at once, and detailed pointcloud information need to be retained in the segmented pointcloud. Future work will involve investigations on potential industrial applications for the model.



\clearpage



\bibliography{example}  

\begin{thebibliography}{31}
\providecommand{\natexlab}[1]{#1}
\providecommand{\url}[1]{\texttt{#1}}
\expandafter\ifx\csname urlstyle\endcsname\relax
  \providecommand{\doi}[1]{doi: #1}\else
  \providecommand{\doi}{doi: \begingroup \urlstyle{rm}\Url}\fi

\bibitem[{Armeni} et~al.(2017){Armeni}, {Sax}, {Zamir}, and
  {Savarese}]{2017arXiv170201105A}
I.~{Armeni}, A.~{Sax}, A.~R. {Zamir}, and S.~{Savarese}.
\newblock {Joint 2D-3D-Semantic Data for Indoor Scene Understanding}.
\newblock \emph{ArXiv e-prints}, Feb. 2017.

\bibitem[Thomas et~al.(2019)Thomas, Qi, Deschaud, Marcotegui, Goulette, and
  Guibas]{kpconv}
H.~Thomas, C.~R. Qi, J.~Deschaud, B.~Marcotegui, F.~Goulette, and L.~J. Guibas.
\newblock Kpconv: Flexible and deformable convolution for point clouds.
\newblock \emph{CoRR}, abs/1904.08889, 2019.
\newblock URL \url{http://arxiv.org/abs/1904.08889}.

\bibitem[Qi et~al.(2016)Qi, Su, Nie{\ss}ner, Dai, Yan, and
  Guibas]{qi2016volumetric}
C.~R. Qi, H.~Su, M.~Nie{\ss}ner, A.~Dai, M.~Yan, and L.~Guibas.
\newblock Volumetric and multi-view cnns for object classification on 3d data.
\newblock In \emph{Proc. Computer Vision and Pattern Recognition (CVPR), IEEE},
  2016.

\bibitem[Li et~al.(2016)Li, Pirk, Su, Qi, and
  Guibas]{DBLP:journals/corr/LiPSQG16}
Y.~Li, S.~Pirk, H.~Su, C.~R. Qi, and L.~J. Guibas.
\newblock {FPNN:} field probing neural networks for 3d data.
\newblock \emph{CoRR}, abs/1605.06240, 2016.
\newblock URL \url{http://arxiv.org/abs/1605.06240}.

\bibitem[Engelcke et~al.(2016)Engelcke, Rao, Wang, Tong, and
  Posner]{DBLP:journals/corr/EngelckeRWTP16}
M.~Engelcke, D.~Rao, D.~Z. Wang, C.~H. Tong, and I.~Posner.
\newblock Vote3deep: Fast object detection in 3d point clouds using efficient
  convolutional neural networks.
\newblock \emph{CoRR}, abs/1609.06666, 2016.
\newblock URL \url{http://arxiv.org/abs/1609.06666}.

\bibitem[Qi et~al.(2016)Qi, Su, Mo, and Guibas]{pointnet}
C.~R. Qi, H.~Su, K.~Mo, and L.~J. Guibas.
\newblock Pointnet: Deep learning on point sets for 3d classification and
  segmentation.
\newblock \emph{CoRR}, abs/1612.00593, 2016.
\newblock URL \url{http://arxiv.org/abs/1612.00593}.

\bibitem[Zhang et~al.(2019)Zhang, Hua, and Yeung]{shelnet}
Z.~Zhang, B.-S. Hua, and S.-K. Yeung.
\newblock Shellnet: Efficient point cloud convolutional neural networks using
  concentric shells statistics.
\newblock In \emph{Proceedings of the IEEE/CVF International Conference on
  Computer Vision (ICCV)}, October 2019.

\bibitem[Masci et~al.(2015)Masci, Boscaini, Bronstein, and
  Vandergheynst]{DBLP:journals/corr/MasciBBV15}
J.~Masci, D.~Boscaini, M.~M. Bronstein, and P.~Vandergheynst.
\newblock Shapenet: Convolutional neural networks on non-euclidean manifolds.
\newblock \emph{CoRR}, abs/1501.06297, 2015.
\newblock URL \url{http://arxiv.org/abs/1501.06297}.

\bibitem[Xu et~al.(2018)Xu, Fan, Xu, Zeng, and
  Qiao]{10.1007/978-3-030-01237-3_6}
Y.~Xu, T.~Fan, M.~Xu, L.~Zeng, and Y.~Qiao.
\newblock Spidercnn: Deep learning on point sets with parameterized
  convolutional filters.
\newblock In V.~Ferrari, M.~Hebert, C.~Sminchisescu, and Y.~Weiss, editors,
  \emph{Computer Vision -- ECCV 2018}, pages 90--105, Cham, 2018. Springer
  International Publishing.

\bibitem[Qi et~al.(2017)Qi, Yi, Su, and Guibas]{pointnet2}
C.~R. Qi, L.~Yi, H.~Su, and L.~J. Guibas.
\newblock Pointnet++: Deep hierarchical feature learning on point sets in a
  metric space.
\newblock \emph{CoRR}, abs/1706.02413, 2017.
\newblock URL \url{http://arxiv.org/abs/1706.02413}.

\bibitem[Li et~al.(2018)Li, Bu, Sun, and Chen]{pointcnn}
Y.~Li, R.~Bu, M.~Sun, and B.~Chen.
\newblock Pointcnn.
\newblock \emph{CoRR}, abs/1801.07791, 2018.
\newblock URL \url{http://arxiv.org/abs/1801.07791}.

\bibitem[Zhao et~al.(2019)Zhao, Jiang, Fu, and Jia]{pointweb}
H.~Zhao, L.~Jiang, C.-W. Fu, and J.~Jia.
\newblock Pointweb: Enhancing local neighborhood features for point cloud
  processing.
\newblock In \emph{Proceedings of the IEEE/CVF Conference on Computer Vision
  and Pattern Recognition (CVPR)}, June 2019.

\bibitem[Komarichev et~al.(2019)Komarichev, Zhong, and Hua]{acnn}
A.~Komarichev, Z.~Zhong, and J.~Hua.
\newblock {A-CNN:} annularly convolutional neural networks on point clouds.
\newblock \emph{CoRR}, abs/1904.08017, 2019.
\newblock URL \url{http://arxiv.org/abs/1904.08017}.

\bibitem[Mao et~al.(2019)Mao, Wang, and Li]{interpolate}
J.~Mao, X.~Wang, and H.~Li.
\newblock Interpolated convolutional networks for 3d point cloud understanding.
\newblock In \emph{Proceedings of the IEEE/CVF International Conference on
  Computer Vision (ICCV)}, October 2019.

\bibitem[Landrieu and Simonovsky(2017)]{DBLP:journals/corr/abs-1711-09869}
L.~Landrieu and M.~Simonovsky.
\newblock Large-scale point cloud semantic segmentation with superpoint graphs.
\newblock \emph{CoRR}, abs/1711.09869, 2017.
\newblock URL \url{http://arxiv.org/abs/1711.09869}.

\bibitem[Tang et~al.(2019)Tang, Li, and Yu]{Tang2019ChebNetEA}
S.~Tang, B.~Li, and H.~Yu.
\newblock Chebnet: Efficient and stable constructions of deep neural networks
  with rectified power units using chebyshev approximations.
\newblock \emph{ArXiv}, abs/1911.05467, 2019.

\bibitem[Kipf and Welling(2016)]{DBLP:journals/corr/KipfW16}
T.~N. Kipf and M.~Welling.
\newblock Semi-supervised classification with graph convolutional networks.
\newblock \emph{CoRR}, abs/1609.02907, 2016.
\newblock URL \url{http://arxiv.org/abs/1609.02907}.

\bibitem[Hamilton et~al.(2017)Hamilton, Ying, and
  Leskovec]{DBLP:journals/corr/HamiltonYL17}
W.~L. Hamilton, R.~Ying, and J.~Leskovec.
\newblock Inductive representation learning on large graphs.
\newblock \emph{CoRR}, abs/1706.02216, 2017.
\newblock URL \url{http://arxiv.org/abs/1706.02216}.

\bibitem[Veli{\v{c}}kovi{\'{c}} et~al.(2018)Veli{\v{c}}kovi{\'{c}}, Cucurull,
  Casanova, Romero, Li{\`{o}}, and Bengio]{velickovic2018graph}
P.~Veli{\v{c}}kovi{\'{c}}, G.~Cucurull, A.~Casanova, A.~Romero, P.~Li{\`{o}},
  and Y.~Bengio.
\newblock {Graph Attention Networks}.
\newblock \emph{International Conference on Learning Representations}, 2018.
\newblock URL \url{https://openreview.net/forum?id=rJXMpikCZ}.

\bibitem[Wang et~al.(2018)Wang, Sun, Liu, Sarma, Bronstein, and
  Solomon]{DBLP:journals/corr/abs-1801-07829}
Y.~Wang, Y.~Sun, Z.~Liu, S.~E. Sarma, M.~M. Bronstein, and J.~M. Solomon.
\newblock Dynamic graph {CNN} for learning on point clouds.
\newblock \emph{CoRR}, abs/1801.07829, 2018.
\newblock URL \url{http://arxiv.org/abs/1801.07829}.

\bibitem[Shen et~al.(2017)Shen, Feng, Yang, and Tian]{mining}
Y.~Shen, C.~Feng, Y.~Yang, and D.~Tian.
\newblock Neighbors do help: Deeply exploiting local structures of point
  clouds.
\newblock \emph{CoRR}, abs/1712.06760, 2017.
\newblock URL \url{http://arxiv.org/abs/1712.06760}.

\bibitem[Wang et~al.(2019)Wang, Huang, Hou, Zhang, and Shan]{Wang2019_GACNet}
L.~Wang, Y.~Huang, Y.~Hou, S.~Zhang, and J.~Shan.
\newblock Graph attention convolution for point cloud semantic segmentation.
\newblock In \emph{The IEEE Conference on Computer Vision and Pattern
  Recognition (CVPR)}, June 2019.

\bibitem[Engelmann et~al.(2018)Engelmann, Kontogianni, Hermans, and Leibe]{eng}
F.~Engelmann, T.~Kontogianni, A.~Hermans, and B.~Leibe.
\newblock Exploring spatial context for 3d semantic segmentation of point
  clouds.
\newblock \emph{CoRR}, abs/1802.01500, 2018.
\newblock URL \url{http://arxiv.org/abs/1802.01500}.

\bibitem[Wang et~al.(2018)Wang, Sun, Liu, Sarma, Bronstein, and Solomon]{dgcnn}
Y.~Wang, Y.~Sun, Z.~Liu, S.~E. Sarma, M.~M. Bronstein, and J.~M. Solomon.
\newblock Dynamic graph {CNN} for learning on point clouds.
\newblock \emph{CoRR}, abs/1801.07829, 2018.
\newblock URL \url{http://arxiv.org/abs/1801.07829}.

\bibitem[Tan et~al.(2019)Tan, Pang, and Le]{tan2019efficientdet}
M.~Tan, R.~Pang, and Q.~V. Le.
\newblock Efficientdet: Scalable and efficient object detection, 2019.

\bibitem[Demantké et~al.(2011)Demantké, Mallet, David, and Vallet]{article}
J.~Demantké, C.~Mallet, N.~David, and B.~Vallet.
\newblock Dimensionality based scale selection in 3d lidar point clouds.
\newblock \emph{Proceedings of the ISPRS Workshop Laser Scanning}, 38:\penalty0
  97--102, 01 2011.
\newblock \doi{10.5194/isprsarchives-XXXVIII-5-W12-97-2011}.

\bibitem[Landrieu and Boussaha(2019)]{DBLP:journals/corr/abs-1904-02113}
L.~Landrieu and M.~Boussaha.
\newblock Point cloud oversegmentation with graph-structured deep metric
  learning.
\newblock \emph{CoRR}, abs/1904.02113, 2019.
\newblock URL \url{http://arxiv.org/abs/1904.02113}.

\bibitem[Gaidon et~al.(2016)Gaidon, Wang, Cabon, and
  Vig]{DBLP:journals/corr/GaidonWCV16}
A.~Gaidon, Q.~Wang, Y.~Cabon, and E.~Vig.
\newblock Virtual worlds as proxy for multi-object tracking analysis.
\newblock \emph{CoRR}, abs/1605.06457, 2016.
\newblock URL \url{http://arxiv.org/abs/1605.06457}.

\bibitem[Hu et~al.(2019)Hu, Yang, Xie, Rosa, Guo, Wang, Trigoni, and
  Markham]{hu2019randla}
Q.~Hu, B.~Yang, L.~Xie, S.~Rosa, Y.~Guo, Z.~Wang, N.~Trigoni, and A.~Markham.
\newblock Randla-net: Efficient semantic segmentation of large-scale point
  clouds.
\newblock \emph{arXiv preprint arXiv:1911.11236}, 2019.

\bibitem[Oono and Suzuki(2019)]{express}
K.~Oono and T.~Suzuki.
\newblock On asymptotic behaviors of graph cnns from dynamical systems
  perspective.
\newblock \emph{CoRR}, abs/1905.10947, 2019.
\newblock URL \url{http://arxiv.org/abs/1905.10947}.

\bibitem[Dehmamy et~al.(2019)Dehmamy, Barab{\'{a}}si, and Yu]{power}
N.~Dehmamy, A.~Barab{\'{a}}si, and R.~Yu.
\newblock Understanding the representation power of graph neural networks in
  learning graph topology.
\newblock \emph{CoRR}, abs/1907.05008, 2019.
\newblock URL \url{http://arxiv.org/abs/1907.05008}.

\end{thebibliography}

\end{document}